\documentclass[sigconf]{acmart}
\usepackage{url}
\usepackage{amsfonts}
\usepackage{bm} 
\usepackage{subfigure}
\usepackage{enumitem}
\usepackage{threeparttable}
\usepackage{multirow}
\usepackage{makecell}
\usepackage{amsmath}
\usepackage{marvosym}
\usepackage{xcolor}

\newcommand{\boldres}[1]{{\textbf{\textcolor{red}{#1}}}}
\newcommand{\secondres}[1]{{\underline{\textcolor{blue}{#1}}}}

\AtBeginDocument{%
  }

\copyrightyear{2024}
\acmYear{2024}
\setcopyright{acmlicensed}\acmConference[CIKM '24]{Proceedings of the 33rd ACM International Conference on Information and Knowledge Management}{October 21--25, 2024}{Boise, ID, USA} \acmBooktitle{Proceedings of the 33rd ACM International Conference on Information and Knowledge Management (CIKM '24), October 21--25, 2024, Boise, ID, USA}
\acmDOI{10.1145/3627673.3680072} \acmISBN{979-8-4007-0436-9/24/10}

\begin{document}

\title{Multiscale Representation Enhanced Temporal Flow Fusion Model for Long-Term Workload Forecasting}


\author{Shiyu Wang}
\authornote{Both authors contributed equally to this research.}
\email{weiming.wsy@antgroup.com}
\orcid{0000-0001-5376-6761}
\affiliation{%
  \institution{Ant Group}
  \city{Hangzhou}
  \state{Zhejiang}
  \country{China}
}

\author{Zhixuan Chu \textsuperscript{\Letter}}
\affiliation{%
  \institution{The State Key Laboratory of Blockchain and Data Security, Zhejiang University}
  \city{Hangzhou}
  \state{Zhejiang}
  \country{China}}
\email{zhixuanchu@zju.edu.cn}

\author{Yinbo Sun}
\authornotemark[1]
\email{yinbo.syb@antgroup.com}
\affiliation{%
  \institution{Ant Group}
  \city{Hangzhou}
  \state{Zhejiang}
  \country{China}
}

\author{Yu Liu}
\email{nuoman.ly@antgroup.com}
\affiliation{%
  \institution{Ant Group}
  \city{Hangzhou}
  \state{Zhejiang}
  \country{China}
}

\author{Yuliang Guo}
\email{yuliang.gyl@antgroup.com}
\affiliation{%
  \institution{Ant Group}
  \city{Hangzhou}
  \state{Zhejiang}
  \country{China}
}

\author{Yang Chen}
\email{chenyang.chenyang@antgroup.com}
\affiliation{%
  \institution{Ant Group}
  \city{Hangzhou}
  \state{Zhejiang}
  \country{China}
}

\author{Huiyang Jian}
\email{jianhuiyang.jhy@antgroup.com}
\affiliation{%
  \institution{Ant Group}
  \city{Hangzhou}
  \state{Zhejiang}
  \country{China}
}

\author{Lintao Ma \textsuperscript{\Letter}}
\email{lintao.mlt@antgroup.com}
\affiliation{%
  \institution{Ant Group}
  \city{Hangzhou}
  \state{Zhejiang}
  \country{China}
}

\author{Xingyu Lu}
\email{sing.lxy@antgroup.com}
\affiliation{%
  \institution{Ant Group}
  \city{Hangzhou}
  \state{Zhejiang}
  \country{China}
}

\author{Jun Zhou}
\email{jun.zhoujun@antgroup.com}
\affiliation{%
  \institution{Ant Group}
  \city{Hangzhou}
  \state{Zhejiang}
  \country{China}
}

\renewcommand{\shortauthors}{Shiyu Wang et al.}

\begin{abstract}
    Accurate workload forecasting is critical for efficient resource management in cloud computing systems, enabling effective scheduling and autoscaling. Despite recent advances with transformer-based forecasting models, challenges remain due to the non-stationary, nonlinear characteristics of workload time series and the long-term dependencies. In particular, inconsistent performance between long-term history and near-term forecasts hinders long-range predictions. This paper proposes a novel framework leveraging self-supervised multiscale representation learning to capture both long-term and near-term workload patterns. The long-term history is encoded through multiscale representations while the near-term observations are modeled via temporal flow fusion. These representations of different scales are fused using an attention mechanism and characterized with normalizing flows to handle non-Gaussian/non-linear distributions of time series. Extensive experiments on 9 benchmarks demonstrate superiority over existing methods. 
\end{abstract}


%

\maketitle

\section{Introduction}
With the continuous expansion of cloud computing, efficient resource management has become a crucial issue for cloud systems \cite{nayak2018nature,yin2014system,deng2019distributed,deng2018establishment}. Accurately predicting future workloads is essential for effective resource scheduling \cite{roy2011efficient,masdari2020survey}. For microservices systems, request per second (RPS) is their main service capacity metric, which quantifies their workload \cite{hua2023kae,wang2023full}. Therefore, we achieve workload forecasting for microservices by predicting RPS.
In recent years, major cloud service providers have successively launched their own resource scaling service frameworks, such as Google's Autopilot \cite{rzadca2020autopilot}, Microsoft's FIRM \cite{qiu2020firm}, and Amazon's AWS Autoscaling, which adopt different workload forecasting methods. Autopilot uses ARIMA \cite{box1968somearima,makridakis1997arma}, FIRM uses historical statistical methods, and AWS adopts DeepAR \cite{salinas2020deepar}, among others. Especially recently, with the increasing complexity of cloud computing services, there has been widespread attention and research on forecasting long-term workloads time series \cite{hua2023kae,chatzopoulos2023atrapos,li2023elastic,feng2023group,wang2023timemixer,liu2023itransformer}.
Notably, a plethora of transformer-based methods, such as Informer \cite{zhou2021informer}, Autoformer \cite{kingma2013auto}, Fedformer \cite{zhou2022fedformer}, and PatchTST \cite{nie2023patchtst}, have emerged in abundance. Notwithstanding, it is imperative to highlight that current research confronts the following challenges due to the complex characteristics of time series:
\begin{itemize}
\item  \textbf{Challenge 1}: \textbf{The multi-periodic, non-stationary, and long-term dependencies.} In long-term forecasting, workloads time series data is usually high-frequency and non-stationary with complex multi-periodic characteristics, such as hourly, daily, and weekly cycles. Furthermore, these data exhibit long-term dependent properties. Hence, capturing the multi-periodic and long-term characteristics of time series data is critical in addressing the challenges of long-term forecasting.

\item \textbf{Challenge 2}: \textbf{The non-Gaussian/non-linear characteristics of workloads time series that have not been adequately characterized.} Time series data often exhibits non-Gaussian/non-linear distributions, which previous studies have predominantly assumed to be Gaussian. This assumption has made it challenging for these studies to adapt to the real distribution of time series data.

\item \textbf{Challenge 3}: \textbf{The inconsistency between the long-term history of time series and near-term observations.} The significant fluctuations and changes in time series data often result in inconsistent performance between long-term history and near-term observations. While long-term history exhibits complex multi-periodic characteristics, near-term observations show a rapidly changing trend. Therefore, integrating the characteristics of both long-term history and near-term observations is crucial for accurate long-term forecasting.
\end{itemize}

Recently, significant progress has been made in representation learning, particularly in the area of self-supervised contrastive learning \cite{he2020momentummoco,chen2020improvedmocov2,chen2021mocov3,chen2020simplesimclr,chen2020bigsimclrv2,chen2021exploringsimsiam}. Harness representation learning to capture the long-term dependency and multi-periodic characteristics of time series data is indeed a potent way. Undoubtedly, this introduces a fresh approach to tackling the aforementioned challenges. Unfettered by the limitations of the original end-to-end model, we can devise a novel learning paradigm for long-term time series forecasting.  

In this work, we propose a two-stage framework that consists of a \textbf{\textit{pretraining representation stage}} and a \textbf{\textit{fusion prediction stage}}. \textbf{During the pretraining representation stage}, we pre-train a multiscale time series representation model using a contrastive learning method in both time and frequency domains. This model extracts representations of different scales from the long-term history of the time series to characterize long-term dependencies and complex multi-periodic patterns \cite{chu2024task}. \textbf{In the fusion prediction stage}, a temporal flow fusion model captures the changing trends in near-term observations. The multiscale representations of the long-term history and the nearby observations are fused through a FusionAttention module employing a multi-head attention mechanism. Furthermore, we use normalizing flow to model the non-Gaussian/non-linear properties of the time series. Our approach produces accurate predictions that capture both long-term patterns and near-term trends. 

Our Contributions can be summarized as follows:
\begin{itemize}
 \item This paper presents a novel long-term workload forecasting framework unifying multiscale time series representation learning and temporal flow fusion modeling to capture both long-term historical patterns and near-term trends. This unified approach leads to superior prediction accuracy.

\item We propose an original multiscale representation method applying contrastive learning in time and frequency domains to encode long-term dependencies and multi-periodic patterns from long-term history.

\item We conducted extensive experiments on nine benchmarks, and our method achieved consistent state-of-the-art performance. Furthermore, we have conducted large-scale deployment in the real-world as a cornerstone of workload forecasting in the Alipay cloud resource management system.
\end{itemize}

\section{Related Work}
\paragraph{\textbf{Time Series (TS) Forecasting}} 
Due to the immense importance of time series forecasting, various models have been well developed.
In recent years, a variety of time series forecasting models, particularly those based on deep learning methods, have become increasingly popular \cite{xue2023prompt,jin2023time,xue2023easytpp,zhou2023ptse,chen2023monotonic,wang2022end,wang2024neuralreconciler,wang2023flow}. These models have introduced many novel structures and have outperformed classical models such as ARIMA and VAR.
Informer \cite{zhou2021informer} is a prob-sparse self-attention mechanism-based model to enhance the prediction capacity in long-sequence TS forecasting.
Autoformer \cite{wu2021autoformer} is a decomposition architecture that incorporates the series decomposition block as an inner operator.
Fedformer \cite{zhou2022fedformer} is a decomposed Transformer architecture that utilizes a mixture of experts for seasonal-trend decomposition and is enhanced with frequency information.
Non-stationary Transformers \cite{liu2022nonstationary} is to enhance the predictability of time series while maximizing the model's predictive capacity.
PatchTST \cite{nie2023patchtst} is an effective design of Transformer-based models for time series forecasting tasks by introducing two key components: patching and channel-independent structure. 

\paragraph{\textbf{Time Series (TS) Representation}}
Representation learning has recently achieved great success in advancing TS research by characterizing the long temporal dependencies and complex periodicity based on the contrastive method \cite{zhang2023self,deldari2021time,hou2021stock,eldele2021time2,zhang2022self}. 
TS2Vec \cite{yue2022ts2vec} was recently proposed as a universal framework for learning TS representations by performing contrastive learning in a hierarchical loss over augmented context views. 
COST \cite{woo2022cost} proposed a new TS representation learning framework for long-sequence TS forecasting, which applies contrastive learning methods to learn disentangled seasonal-trend representations.
TST \cite{zerveas2021transformertst} is a newly developed framework that utilizes the transformer encoder architecture for multivariate time series representation learning. 
LaST \cite{wang2022learninglast} utilizes variational inference to separate seasonal-trend representations in the latent space.
TimeMAE \cite{cheng2023timemae} is a novel self-supervised paradigm for learning transferrable time series representations based on transformer networks

\paragraph{\textbf{Normalizing flow}}
Normalizing flows(NF) \cite{kobyzev2020normalizing}, which learn a distribution by transforming the data to samples from a tractable distribution where both sampling and density estimation can be efficient and exact, have been proven to be powerful density approximations \cite{papamakarios2019normalizing,papamakarios2021normalizing,rezende2015variational}.  NF are invertible neural networks that typically transform isotropic Gaussians to fit a more complex data distribution \cite{kobyzev2020normalizing}.
They  map from  $\mathbb{R}^{D}$ to $\mathbb{R}^{D}$ such that densities $p_{Y}$ on the input space $Y  \in \mathbb{R}^{D}$ are transformed into some tractable distribution $p_{Z}$ (e.g., an isotropic Gaussian) on space $Z  \in \mathbb{R}^{D}$.  This mapping function, $f:Y \rightarrow Z$, and inverse mapping function, $f^{-1}:Z \rightarrow Y$ is composed of a sequence of bijections or invertible functions, and we can express the target distribution densities $p_{Y}(\bm{y})$ by
\begin{equation}
p_{Y}(\bm{y})=p_{Z}(\bm{z})|det(\frac{\partial	f(\bm{y})}{\partial \bm{y}})| \mbox{ ,}\end{equation}
where $\partial f(y)/\partial y$ is the Jacobian of $f$ at $y$.

For mapping function $f$,  we can employ RealNVP \cite{dinh2017densityrealnvp} architecture, which is a neural network composed of a series of parametrized invertible transformations with a lower triangular Jacobian structure and vector component permutations in order to capture complex dependencies. 
It leaves the part of its inputs unchanged and transforms the other part via functions of the un-transformed variables (with superscript denoting the coordinate indices)
\begin{equation}
  \left\{ 
    \begin{array}{lr}   
        y^{1:d}=x^{1:d} \\
        y^{d+1:D}=x^{d+1:D} \odot exp(s(x)^{1:d}+t(x^{1:d})) 
    \end{array}
 \right.  
\mbox{ ,}\end{equation}
where $\odot$ is an element wise product, $s()$ is a scaling and $t()$ a translation function from $\mathbb{R}^{D} \mapsto \mathbb{R}^{D-d}$, using neural networks.

\section{Methodology}
\begin{figure}[h]
  \centering
  \includegraphics[width=\linewidth]{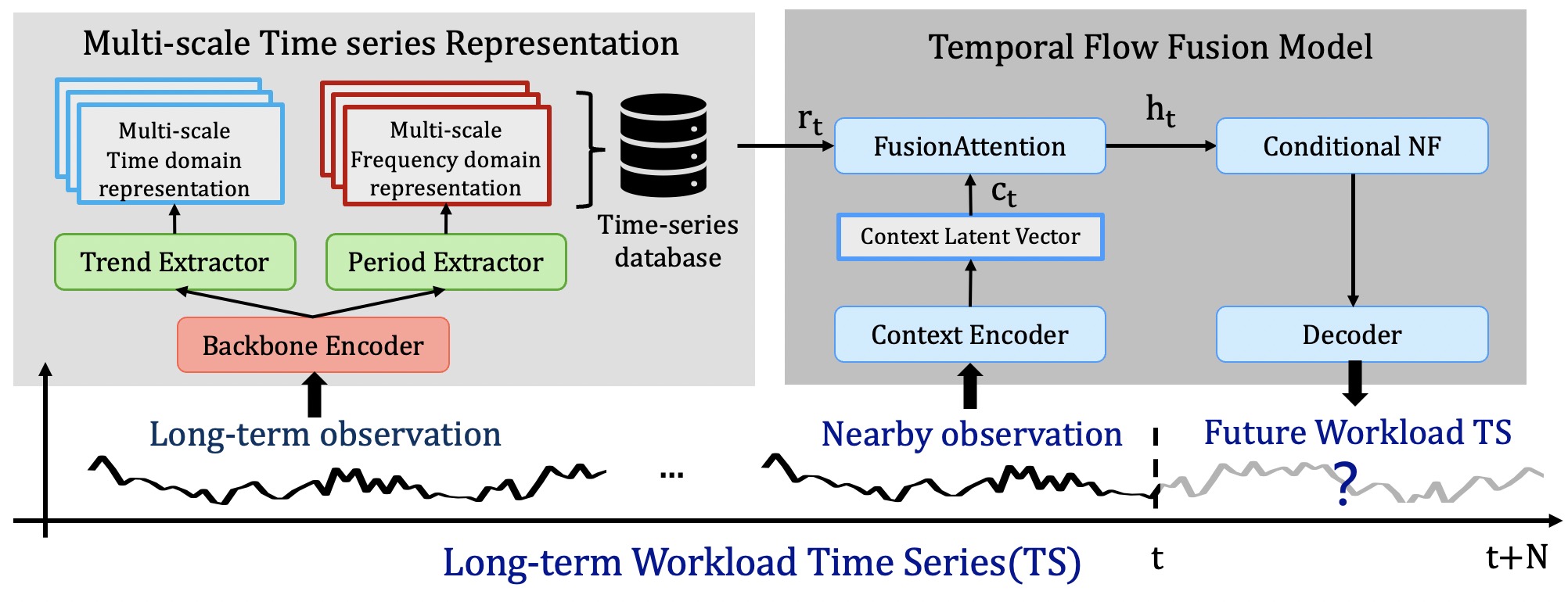}
  \vspace{-3mm}
  \caption{\textbf{Framework Architecture.}}
  \vspace{-3mm}
\end{figure}

Due to the high-frequency and non-stationary nature of the long-term time series (TS), along with their complex long temporal dependencies spanning daily, weekly, monthly, and quarterly periodicities, it becomes challenging to memorize historical data and learn these dependencies for backcasting. In light of this, we propose a method for representing complex historical TS as compressed vectors and storing them in a TS database. To achieve accurate predictions, we design a \textit{Temporal Flow Fusion Model} that integrates these long-term historical TS representations with near-term observations from nearby windows.

\subsection{Multiscale Time Series (TS) Representation}
Given TS $y \in \mathbb{R}^{T \times F}$ with backcast window $h$, our goal is to learn a non-linear embedding function $f_{\theta
}$ that maps $\{y_{t-h}...y_{t}\}$ to its representation $r_{t}=[r^{T}_{t},r^{F}_{t}]$, where $r_{t} \in \mathbb{R}^{K}$ is for each time stamp $t$, $r^{T}_{t} \in \mathbb{R}^{K_{T}}$ is the time domain representation, $r^{F}_{t} \in \mathbb{R}^{K_{F}}$ denotes that of frequency domain and $K=K_{T}+K_{F}$ is the dimension of representation vectors.  In the encoding representation stage, by using backcast windows of various lengths, we can obtain a representation of different scales.

\begin{figure}[h]
  \centering
  \includegraphics[width=\linewidth]{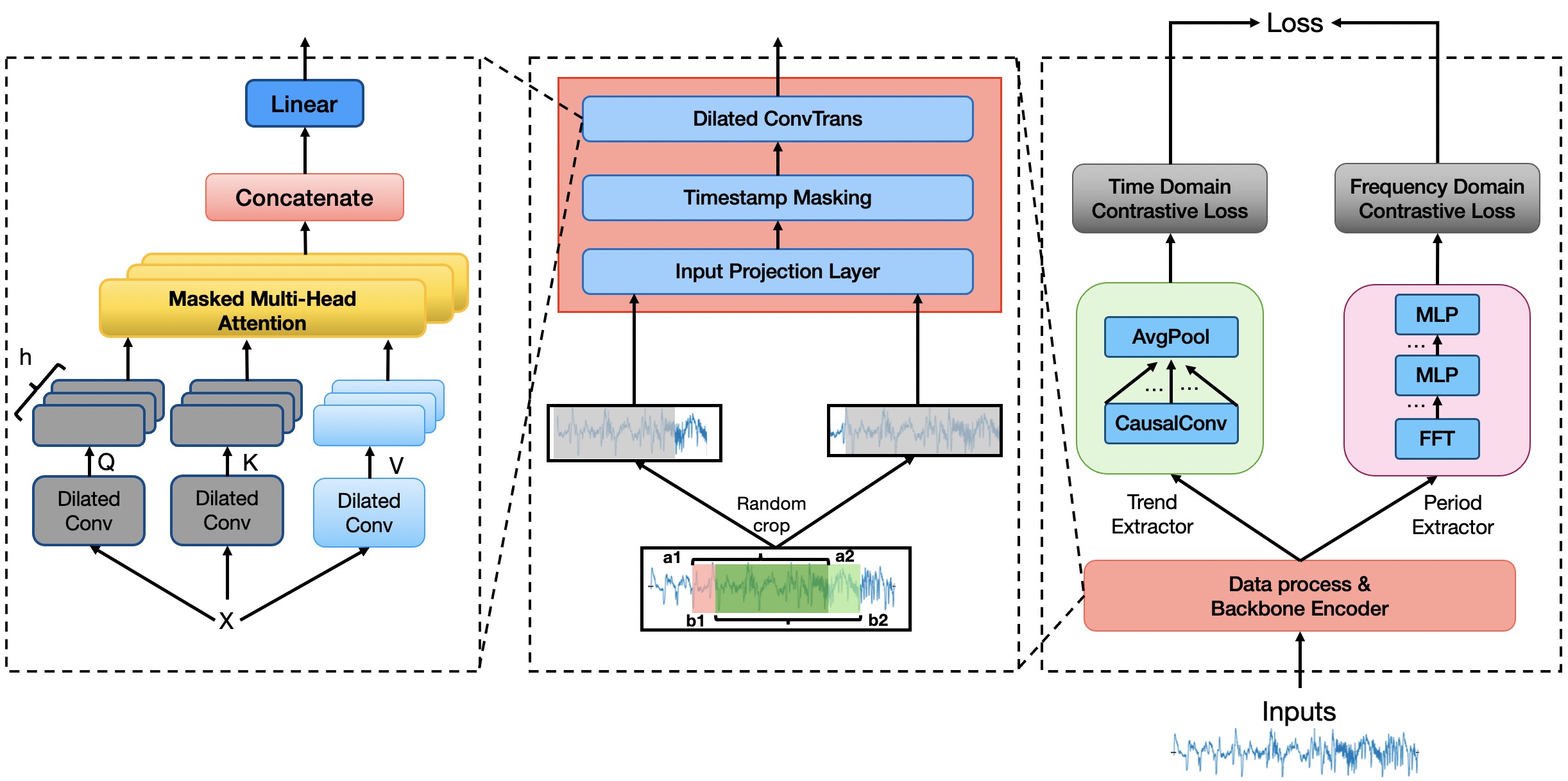}
  \vspace{-5mm}
  \caption{\textbf{Multiscale Time Series Representation}}
  \vspace{-3mm}
\end{figure}

In Figure 2, we can observe the generation of representation using a schematic view. This process initiates with the random sampling of two overlapping subseries from the input time series, which is then followed by individual data augmentation for each subseries. Subsequently, for the input projection layer, we employ Multilayer Perceptron (MLP), and the original input ${y_{t}}$ is mapped into a high-dimensional latent vector ${z_{t}}$.  To generate an augmented context view, we employ timestamp masking to mask latent vectors at randomly selected timestamps. The contextual embeddings at each timestamp are then extracted using the CovnTrans backbone encoder. We further extract trends in the time domain and periods in the frequency domain using CausalConv and Fast Fourier transform (FFT), respectively, from the contextual embeddings. In the end, we carry out contrastive learning in both the time and frequency domains.

In the subsequent sections, we provide a detailed description of each of these components.

\textbf{Random Cropping}  is a popular data augmentation technique used in contrastive learning for generating new context views. We can randomly sample two overlapping time segments $[a_1, a_2]$ and $[b_1, b_2]$ from TS $y \in \mathbb{R}^{T \times F}$ that satisfy $0<a_1<b_1<a_2<b_2$ $\le$ $T$. Note that contextual representations on the overlapped segment $[b1, a2]$ ensure consistency for two context views.

\textbf{Timestamp Masking} aims to produce an augmented context view by randomly masking the timestamps of a TS. We can mask off the latent vector $z=\{z_{t}\}$  after the Input Projection Layer along the time axis with a binary mask $m \in \{0, 1\}^T$, the elements of which are independently sampled from a Bernoulli distribution with $p = 0.5$. 

\textbf{Backbone Encoder} is used to extract the contextual representation at each timestamp. Our selected Backbone Encoder is the 1-layer causal convolution Transformer (ConvTrans), which leverages convolutional multi-head self-attention to capture long- and short-term dependencies.

Specifically, given TS $y \in \mathbb{R}^{T \times F}$, ConvTrans transforms $y$ (as input) into dimension $l$ via dilated causal convolution layer as follows:
\begin{equation*}
\begin{aligned}
Q&=Dilated Conv(y) \\
K&=Dilated Conv(y) \\
V&=Dilated Conv(y) 
\end{aligned}\mbox{ ,}
\end{equation*}
where $\bm{Q} \in \mathbb{R}^{dl \times dh}$, $\bm{K} \in \mathbb{R}^{dl \times dh}$, and $\bm{V} \in \mathbb{R}^{dl \times dh}$ (we denote the length of time steps as $dh$). 
After these transformations, the scaled dot-product attention computes the sequence of vector outputs via:
\begin{equation*}
    {S}=\text{Attention}({Q},\ {K},\ {V})=\operatorname{softmax}\left({{Q} {K}^{T}}/{\sqrt{d_{K}}} \cdot {M}\right ) {V}
 \mbox{ ,}
 \end{equation*}
 where the mask matrix $M$ can be applied to filter out right-ward attention (or future information leakage) by setting its upper-triangular elements to $-\infty$ and normalization factor $d_{K}$ is the dimension of $W_{h}^{K}$ matrix.  Finally, all outputs $S$ are concatenated and linearly projected again into the next layer. After the above series of operations, we use this backbone $f_{\theta}$ to extract the contextual embedding (intermediate representations ) at each timestamp as $\tilde{r}=f_{\theta}(y)$.

\paragraph{\textbf{Time Domain Contrastive Learning}} A straightforward approach for extracting the underlying trend of a time series (TS) is to employ a collection of 1d causal convolution layers (CasualConv) with varying kernel sizes, along with an average-pooling operation to generate the representations, as shown below:
\begin{equation*}
\tilde{r}^{(T,i)}=CausalConv(\tilde{r},2^i) 
\end{equation*}
\begin{equation*}
r^{T}=AvgPool(\tilde{r}^{(T,1)},\tilde{r}^{(T,2)}...,\tilde{r}^{(T,L)}) \mbox{ ,}
\end{equation*}
where $L$ is a hyper-parameter denoting the number of CasualConv, $2^i$ ($i=0,..., L$) is the kernel size of each  CasualConv, $\tilde{r}$ is above intermediate representations from the backbone encoder, followed by average-pool over the $L$ representations to obtain time-domain representation $\tilde{r}^{(T)}$. To learn discriminative representations over time, we use the time domain contrastive loss, which takes the representations at the same timestamp from two views of the input TS as positive samples ($r^{T}_{i,t},\hat{r}^{T}_{i,t}$), while those at different timestamps from the same time series as negative samples, formulated as
\begin{small}
\begin{equation*}
\mathcal{L}_{time}=-log\frac{exp(r^{T}_{i,t}\cdot\hat{r}^{T}_{i,t})}
{\sum_{t^{\prime} \in \mathcal{T}}(exp(r^{T}_{i,t}\cdot\hat{r}^{T}_{i,t^{\prime}})+\mathbb{I}(t \ne t^{\prime})exp(r^{T}_{i,t}\cdot r^{T}_{i,t^{\prime}})}
\end{equation*}
\end{small}
, where $\mathcal{T}$ is the set of timestamps within the overlap of the two subseries, subscript $i$ is the index of the input TS sample, and $t$ is the timestamp.

\paragraph{\textbf{Frequency Domain Contrastive Learning}}
Since spectral analysis has proven to be effective in detecting periods, we utilize Fast Fourier Transforms (FFT) to convert the intermediate representations mentioned above to the frequency domain. This allows us to identify various periodic patterns. By combining the FFT and MLP techniques, we can create a period extractor that extracts the frequency spectrum from the contextual embedding and translates it into the freq-based representation $r^{F}_{t}$.

We implement the frequency domain contrastive loss with an index of $(i,t)$ across the batch instances to train the representations to distinguish between various periodic patterns. The formula for this loss is as follows:
\begin{small}
\begin{equation*}
\mathcal{L}_{Freq}=-log
\frac{exp(r^{F}_{i,t}\cdot\hat{r}^{F}_{i,t})}
{\sum_{j \in \mathcal{D}}(exp(r^{F}_{i,t}\cdot\hat{r}^{F}_{i,t^{\prime}})+\mathbb{I}(i \ne j)exp(r^{F}_{i,t}\cdot r^{F}_{i,t^{\prime}})}
\end{equation*}
\end{small}
, where $\mathcal{D}$ is defined as a batch of TS. We use freq-based representations of other TS at timestamp $t$ in the same batch as negative samples.

The contrastive loss is composed of two losses that are complementary to each other and is defined as
\begin{equation}
\begin{aligned}
      \mathcal{L}=
      \frac{1}{|\mathcal D|T}(\mathcal{L}_{time}+\mathcal{L}_{Freq})
\end{aligned}
 \mbox{ ,}\end{equation}
where $\mathcal D$ denotes a batch of TS.
As previously noted, we pre-train our TS representation model. \textbf{During the encoding representation phase, we use backcast windows of varying lengths to produce representations at different scales.} In this study, we apply this method to encode high-frequency TS data to generate long-term historical TS representations at daily, weekly, monthly, and quarterly intervals.

\subsection{Temporal Flow Fusion Model}
In this section, we will provide a detailed overview of the \textit{Temporal Flow Fusion Model}. The values of TS will be denoted as $y_{t} \in \mathbb{R}$, where $t$ represents the time index within the horizon of $t \in {1,2,...,T}$. It should be noted that we define $x_{t}$ as covariates that are known in the future, such as time features and ID features, at time step $t$.

\begin{figure}[h]
  \centering
  \includegraphics[width=\linewidth]{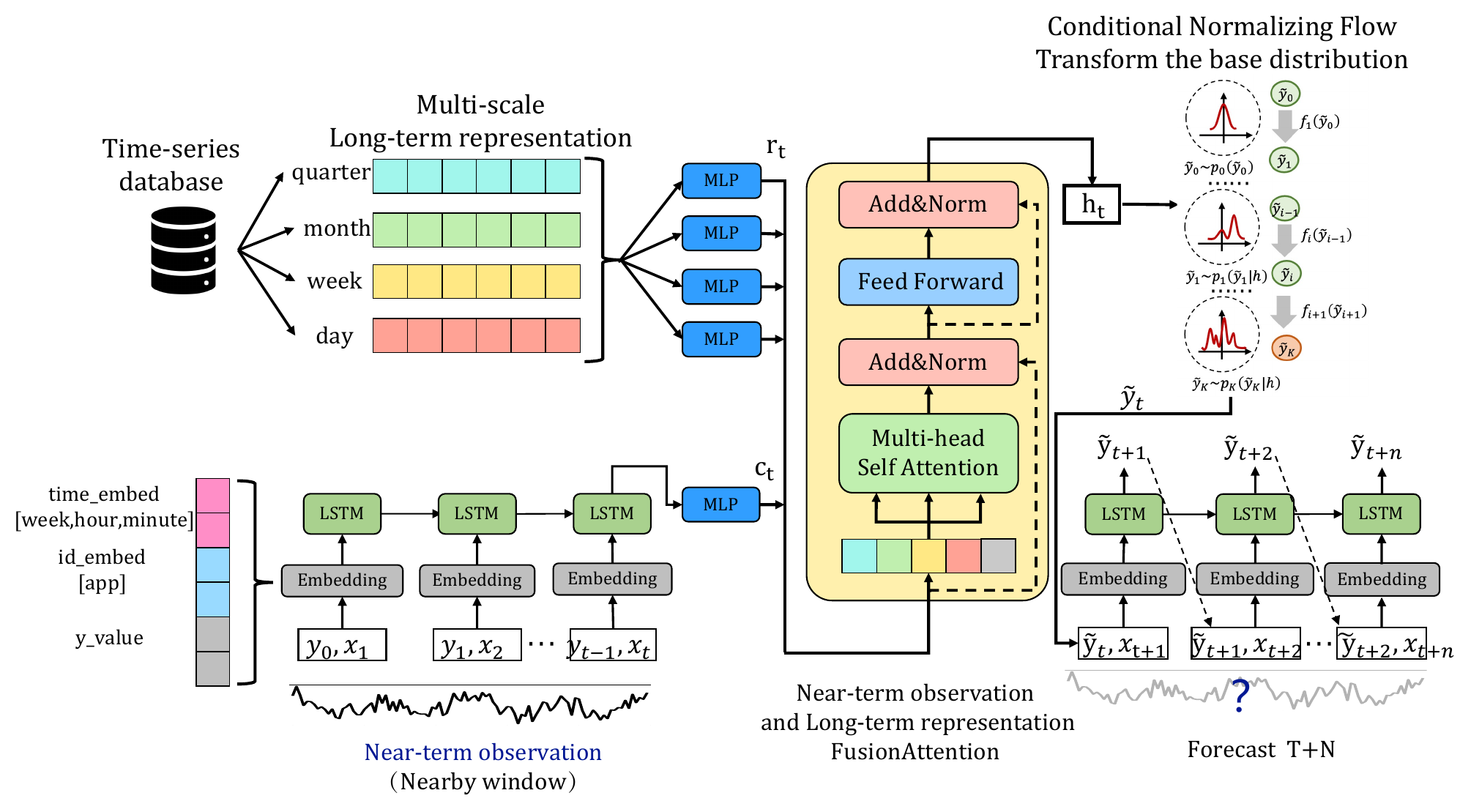}
  \vspace{-5mm}
  \caption{\textbf{Temporal Flow Fusion Model Architecture.}}
  \vspace{-6mm}
\end{figure}

Given the last $L$ observations $y_{t-L},...,y_{t}$, the TS forecasting task aims to predict the future N observations $y_{t+1},...,y_{t+N}$. 
We use $r_t$, the representation of the long-term historical TS, and $h_{t}$ as the context latent vector from near-term observations (nearby window), to predict future observations. Specifically, our Temporal Flow Fusion Model is structured into three steps, as follows:

\textbf{First, loading the TS representation and extracting near-term observation features.} To begin, we load the Multiscale TS representation $r_{t}$ from the TS database, which includes daily, weekly, monthly, and quarterly representations. These representations capture various periods and complex long-term temporal dependencies. Next, we apply Recurrent Neural Networks (RNN) to encode short-term observations (within a near-term window) into a context latent vector $c_{t}$ following Equation 4. This vector captures the nearby window changing pattern of long-term time series data.

\begin{equation}
    \begin{aligned}
        \bm{c_{t}}=RNN(\bm{y}_{t-L:t},\bm{x}_{t-L:t};\Theta) \mbox{ ,}
    \end{aligned}
\end{equation}
where $\Theta$ is the learnable parameters of the RNN.

\textbf{Second, fusing the long-term representation with the near-term observation.} 
After integrating the Multiscale TS representations $r_{t}$ and the context latent vector $c_{t}$ into the same dimension using MLP, we construct the FusionAttention employing a multi-head self-attention module, inspired by the transformer architecture, to fuse the long-term historical TS representations and the near-term observations within the nearby window. This fusion module enhances our ability to make accurate predictions by capturing both long-term dependencies and short-term changes within a day. 

\begin{equation}
    \begin{aligned}
        \bm{h_{t}}=FusionAttention(\bm{c}_{t},\bm{r}_{t};\Phi) \mbox{ ,}
    \end{aligned}
\end{equation}
where $\Phi$ is the learnable parameters of the FusionAttention.

\textbf{Third, employing conditional NF to generate the distribution of the future TS and make predictions.}
Ultimately, we employ the conditional normalizing flow to approximate the probability density of the TS data. Subsequently, we utilize an RNN as the decoder to facilitate autoregressive decoding and accomplish multi-step prediction.

To estimate the probability density of data (in order to obtain a probabilistic forecast), one straightforward method is to use parameterized Gaussian distribution, but as mentioned above, the real-world hierarchical TS data are mostly non-Gaussian/non-linear. Equipped with the powerful density approximator, NF, we are able to tackle this challenge, capturing the nonlinear relationships among the TS data. 

Note that we extend the traditional normalizing flow (\textbf{NF}) to a conditional normalizing flow (\textbf{CNF}). Specifically, we still adopt the Real-NVP architecture,  but we extend Equation 2 by concatenating condition $\bm{h}_{t}$ to both the inputs of the scaling and translation function approximators of the coupling layers as follows:

It should be noted that we expand the conventional normalizing flow to a conditional normalizing flow. More specifically, we maintain the Real-NVP architecture while augmenting Equation 2 through the concatenation of condition $\bm{h}_{t}$ to the inputs of the scaling and translation function approximators of the coupling layers, as following below:
\begin{equation}
  \left\{ 
    \begin{array}{lr}   
        \bm{y}^{1:d}=\bm{z}^{1:d} \\
        \bm{y}^{d+1:D}=\bm{z}^{d+1:D} \odot exp(s(\bm{z}^{1:d},\bm{h})+t(\bm{z}^{1:d},\bm{h})) 
    \end{array}
 \right.  
\mbox{ ,}\end{equation}
where $\bm{z}$ is a noise vector sampled from an isotropic Gaussian, functions $s$ (scale) and $t$ (translation) are usually deep neural networks, which as mentioned above, do not need to be invertible.

To obtain an expressive distribution representation, we can stack \textit{K} layers of conditional flow modules (Real-NVP), generating the conditional distribution of the future sequences of all TS, given the past time $t \in [t-L, t)$.  Specifically, it can be written as a product of factors (as an autoregressive model):
\begin{equation}
p(\bm{\tilde{y}}_{t:T}|\bm{y}_{t-L:t},\bm{x}_{t-L:t};\theta,\phi,\psi)=\prod_{t=t+1}^{T}p(\bm{\tilde{y}}_{t}|\bm{h}_{t}; \Psi)
\mbox{ ,}\end{equation}
where $\bm{\tilde{y}}_{t}$ is the future predictions and $\Psi$ is the parameter of conditional NF.

\textbf{In the training}, given $\mathcal D$, defined as a batch of TS $Y:=\{y_{1},y_{2},...,y_{T}\}$, the representation of the long-term historical TS as $r_{t}$, 
and the associated covariates $X:=\{x_{1},x_{2},...,x_{T}\}$, we can derive the likelihood as:
\begin{equation}
\begin{aligned}
      \mathcal{L}=
      \frac{1}{|\mathcal D|T}\prod_{x1:T,y1:T \in \mathcal D}\prod_{t=1}^{T} p(y_{t}|y_{1:t};x_{1:t}, r_{t},\Theta, \Phi,\Psi) 
\end{aligned}
 \mbox{ ,}\end{equation}
 where $\Theta$ is the learnable parameters of the RNN  and $\Phi$ is the parameter of FusionAttention, and $\psi$ is the parameter of contitional NF.

\section{Experiments}
We conduct extensive experiments to evaluate the performance of our method on long-term forecasting including 9 real-world benchmarks including 8 well-known datasets in time series and a large real-world workload dataset (RPS data of 1589 microservices from Alipay). Furthermore, We have conducted thorough comparisons with 10 well-acknowledged and advanced baselines.

\subsection{Benchmarks}
For long-term forecasting, we conduct the experiments on 9 well-established benchmarks: ETT datasets (including 4 subsets: ETTh1, ETTh2, ETTm1, ETTm2) \cite{zhou2021informer}, Solar-Energy\cite{lai2018modelinglstnet}, Weather, Electricity, and Traffic \cite{wu2021autoformer}, as well as the large \href{https://github.com/traas-stack/forecasting}{Service-Workload datasets}.

\subsection{Baselines}
We compared our method with $10$ advanced baselines. These methods can be divided into two categories: end-to-end forecasting models (including PatchTST \cite{nie2023patchtst}, FEDformer \cite{zhou2022fedformer}, Non-stationary Transformer \cite{liu2022nonstationary}, Autoformer \cite{wu2021autoformer}, Informer \cite{zhou2021informer}) and time series representation models (including TimeMAE \cite{cheng2023timemae}, LaST \cite{wang2022learninglast}, TST\cite{zerveas2021transformertst}, COST \cite{woo2022cost}, TS2VEC \cite{yue2022ts2vec}). Regarding metrics, we utilize the mean square error (MSE) and mean absolute error (MAE) for long-term forecasting. 
\vspace{-2mm}
\begin{align*} \label{equ:metrics}
    \text{MSE} &= (\sum_{i=1}^F (\mathbf{X}_{i} - \widehat{\mathbf{X}}_{i})^2)^{\frac{1}{2}},
    &
    \text{MAE} &= \sum_{i=1}^F|\mathbf{X}_{i} - \widehat{\mathbf{X}}_{i}|,\\
\end{align*}
where $s$ is the periodicity of the data. $\mathbf{X},\widehat{\mathbf{X}}\in\mathbb{R}^{F\times C}$ are the ground truth and prediction results of the future with $F$ time pints and $C$ dimensions. $\mathbf{X}_{i}$ means the $i$-th future time point.

\begin{table*}[tbp]
\vspace{-3mm}
  \caption{Long-term forecasting results on 9 benchmarks. All the results are averaged from 4 different prediction lengths, that is $\{96,192,336,720\}$. A lower MSE or MAE indicates a better prediction.}\label{tab:long_term_forecasting_results}
  \vspace{-3mm}
  \centering
  \begin{threeparttable}
  \begin{small}
  \renewcommand{\multirowsetup}{\centering}
  \setlength{\tabcolsep}{1pt}
  \begin{tabular}{c|cc|cc|cc|cc|cc|cc|cc|cc|cc|cc|cc}
    \toprule
    \multicolumn{1}{c}{\multirow{2}{*}{Models}} &
    \multicolumn{2}{c}{\rotatebox{0}{\scalebox{0.8}{\textbf{Ours}}}} &
    \multicolumn{2}{c}{\rotatebox{0}{\scalebox{0.8}{TimeMAE}}} &
    \multicolumn{2}{c}{\rotatebox{0}{\scalebox{0.8}{TST}}} &
    \multicolumn{2}{c}{\rotatebox{0}{\scalebox{0.8}{LaST}}} &
    \multicolumn{2}{c}{\rotatebox{0}{\scalebox{0.8}{COST}}} &
    \multicolumn{2}{c}{\rotatebox{0}{\scalebox{0.8}{TS2VEC}}} & 
    \multicolumn{2}{c}{\rotatebox{0}{\scalebox{0.8}{PatchTST}}} &
    \multicolumn{2}{c}{\rotatebox{0}{\scalebox{0.8}{FEDformer}}} & 
    \multicolumn{2}{c}{\rotatebox{0}{\scalebox{0.8}{Stationary}}} & 
    \multicolumn{2}{c}{\rotatebox{0}{\scalebox{0.8}{Autoformer}}} &
    \multicolumn{2}{c}{\rotatebox{0}{\scalebox{0.8}{Informer}}} \\
    \multicolumn{1}{c}{} & \multicolumn{2}{c}{\scalebox{0.8}{(\textbf{2023})}} &
    \multicolumn{2}{c}{\scalebox{0.8}{{2023}}}  &
    \multicolumn{2}{c}{\scalebox{0.8}{{2021}}} &
    \multicolumn{2}{c}{\scalebox{0.8}{{2022}}} &
    \multicolumn{2}{c}{\scalebox{0.8}{{2022}}}  &
    \multicolumn{2}{c}{\scalebox{0.8}{{2022}}}  &
    \multicolumn{2}{c}{\scalebox{0.8}{{2023}}}  &
    \multicolumn{2}{c}{\scalebox{0.8}{{2022}}}  &
    \multicolumn{2}{c}{\scalebox{0.8}{{2022}}}  &
    \multicolumn{2}{c}{\scalebox{0.8}{{2021}}}  &
    \multicolumn{2}{c}{\scalebox{0.8}{{2020}}}

    \\
    \cmidrule(lr){2-3} \cmidrule(lr){4-5}\cmidrule(lr){6-7} \cmidrule(lr){8-9}\cmidrule(lr){10-11}\cmidrule(lr){12-13}\cmidrule(lr){14-15}\cmidrule(lr){16-17}\cmidrule(lr){18-19} \cmidrule(lr){20-21} \cmidrule(lr){22-23}
    \multicolumn{1}{c|}{Metric} & \scalebox{0.80}{MSE} & \scalebox{0.80}{MAE} & \scalebox{0.80}{MSE} & \scalebox{0.80}{MAE} & \scalebox{0.80}{MSE} & \scalebox{0.80}{MAE} & \scalebox{0.80}{MSE} & \scalebox{0.80}{MAE} & \scalebox{0.80}{MSE} & \scalebox{0.80}{MAE} & \scalebox{0.80}{MSE} & \scalebox{0.80}{MAE} & \scalebox{0.80}{MSE} & \scalebox{0.80}{MAE} & \scalebox{0.80}{MSE} & \scalebox{0.80}{MAE} & \scalebox{0.80}{MSE} & \scalebox{0.80}{MAE} & \scalebox{0.80}{MSE} & \scalebox{0.80}{MAE} & \scalebox{0.80}{MSE} & \scalebox{0.80}{MAE}\\
    \toprule

    \scalebox{1}{Weather}
    &\boldres{\scalebox{0.80}{0.227}}&\boldres{\scalebox{0.80}{0.263}}&\secondres{\scalebox{0.80}{0.230}}&\secondres{\scalebox{0.80}{0.265}}&\scalebox{0.80}{0.239}&\scalebox{0.80}{0.276}&\scalebox{0.80}{0.232}&\scalebox{0.80}{0.261}&\scalebox{0.80}{0.324}&\scalebox{0.80}{0.329}&\scalebox{0.80}{0.233}&\scalebox{0.80}{0.267}&\scalebox{0.80}{0.275}&\scalebox{0.80}{0.280}&\scalebox{0.80}{0.309}&\scalebox{0.80}{0.360}&\scalebox{0.80}{0.288}&\scalebox{0.80}{0.314}&\scalebox{0.80}{0.338}&\scalebox{0.80}{0.382}&\scalebox{0.80}{0.634}&\scalebox{0.80}{0.548}\\
    \midrule

    \scalebox{1}{Electricity} 
    &\boldres{\scalebox{0.80}{0.160}}&\boldres{\scalebox{0.80}{0.255}}&\scalebox{0.80}{0.205}&\scalebox{0.80}{0.296}&\scalebox{0.80}{0.209}&\scalebox{0.80}{0.289}&\secondres{\scalebox{0.80}{0.186}}&\secondres{\scalebox{0.80}{0.274}}&\scalebox{0.80}{0.215}&\scalebox{0.80}{0.295}&\scalebox{0.80}{0.213}&\scalebox{0.80}{0.293}&\scalebox{0.80}{0.216}&\scalebox{0.80}{0.318} &\scalebox{0.80}{0.207} &\scalebox{0.80}{0.321} &\scalebox{0.80}{0.213} & \scalebox{0.80}{0.327} &\scalebox{0.80}{0.227} & \scalebox{0.80}{0.338}&\scalebox{0.80}{0.311} & \scalebox{0.80}{0.397}\\
    \midrule

    \scalebox{1}{Solar-Energy}
     &\boldres{\scalebox{0.80}{0.198}}&\boldres{\scalebox{0.80}{0.252}}&\secondres{\scalebox{0.80}{0.291}}&\secondres{\scalebox{0.80}{0.308}}&\scalebox{0.80}{0.294}&\scalebox{0.80}{0.310}&\scalebox{0.80}{0.301}&\scalebox{0.80}{0.332}&\scalebox{0.80}{0.320}&\scalebox{0.80}{0.328}&\scalebox{0.80}{0.369}&\scalebox{0.80}{0.361}&\scalebox{0.80}{0.329}&\scalebox{0.80}{0.400}&\scalebox{0.80}{0.243}&\scalebox{0.80}{0.350}&\scalebox{0.80}{0.340}&\scalebox{0.80}{0.380}&\scalebox{0.80}{0.593}&\scalebox{0.80}{0.557}&\scalebox{0.80}{0.231}&\scalebox{0.80}{0.273}\\
    \midrule

    \scalebox{1}{Trafﬁc}
   &\boldres{\scalebox{0.80}{0.391}}&\boldres{\scalebox{0.80}{0.272}}&\secondres{\scalebox{0.80}{0.475}}&\secondres{\scalebox{0.80}{0.310}}&\scalebox{0.80}{0.586}&\scalebox{0.80}{0.362}&\scalebox{0.80}{0.713}&\scalebox{0.80}{0.397}&\scalebox{0.80}{0.435}&\scalebox{0.80}{0.362}&\scalebox{0.80}{0.470}&\scalebox{0.80}{0.350}&\scalebox{0.80}{0.488}&\scalebox{0.80}{0.327}&\scalebox{0.80}{0.609}&\scalebox{0.80}{0.376}&\scalebox{0.80}{0.624}&\scalebox{0.80}{0.340}&\scalebox{0.80}{0.628}&\scalebox{0.80}{0.379}&\scalebox{0.80}{0.764}&\scalebox{0.80}{0.415}\\
    \midrule

    \scalebox{1}{Workload}
   &\boldres{\scalebox{0.80}{0.301}}&\boldres{\scalebox{0.80}{0.290}}&\secondres{\scalebox{0.80}{0.405}}&\secondres{\scalebox{0.80}{0.404}}&\scalebox{0.80}{0.533}&\scalebox{0.80}{0.515}&\scalebox{0.80}{0.722}&\scalebox{0.80}{0.675}&\scalebox{0.80}{0.683}&\scalebox{0.80}{0.611}&\scalebox{0.80}{0.552}&\scalebox{0.80}{0.513}&\scalebox{0.80}{0.591}&\scalebox{0.80}{0.517}&\scalebox{0.80}{0.609}&\scalebox{0.80}{0.536}&\scalebox{0.80}{0.532}&\scalebox{0.80}{0.550}&\scalebox{0.80}{0.557}&\scalebox{0.80}{0.561}&\scalebox{0.80}{0.564}&\scalebox{0.80}{0.553}\\
    \midrule

    \scalebox{1}{ETTh1}
    &\boldres{\scalebox{0.80}{0.401}}&\boldres{\scalebox{0.80}{0.425}}&\secondres{\scalebox{0.80}{0.423}}&\scalebox{0.80}{0.446}&\scalebox{0.80}{0.466}&\scalebox{0.80}{0.462}&\scalebox{0.80}{0.474}&\scalebox{0.80}{0.461}&\scalebox{0.80}{0.485}&\scalebox{0.80}{0.472}&\scalebox{0.80}{0.446}&\scalebox{0.80}{0.456}&\scalebox{0.80}{0.455}&\secondres{\scalebox{0.80}{0.444}}&\scalebox{0.80}{0.440}&\scalebox{0.80}{0.460}&\scalebox{0.80}{0.57}&\scalebox{0.80}{0.536}&\scalebox{0.80}{0.496}&\scalebox{0.80}{0.487}&\scalebox{0.80}{0.840}&\scalebox{0.80}{0.795}\\
    \midrule

    \scalebox{1}{ETTh2}
    &\boldres{\scalebox{0.80}{0.328}}&\secondres{\scalebox{0.80}{0.391}}&\secondres{\scalebox{0.80}{0.380}}&\boldres{\scalebox{0.80}{0.386}}&\scalebox{0.80}{0.404}&\scalebox{0.80}{0.421}&\scalebox{0.80}{0.499}&\scalebox{0.80}{0.497}&\scalebox{0.80}{0.399}&\scalebox{0.80}{0.427}&\scalebox{0.80}{0.417}&\scalebox{0.80}{0.468}&\scalebox{0.80}{0.384}&\scalebox{0.80}{0.406}&\scalebox{0.80}{0.433}&\scalebox{0.80}{0.447}&\scalebox{0.80}{0.526}&\scalebox{0.80}{0.516}&\scalebox{0.80}{0.453}&\scalebox{0.80}{0.462}&\scalebox{0.80}{4.431}&\scalebox{0.80}{1.729}\\
    \midrule

    \scalebox{1}{ETTm1}
    &\boldres{\scalebox{0.80}{0.332}}&\boldres{\scalebox{0.80}{0.380}}&\secondres{\scalebox{0.80}{0.366}}&\scalebox{0.80}{0.391}&\scalebox{0.80}{0.373}&\secondres{\scalebox{0.80}{0.389}}&\scalebox{0.80}{0.398}&\scalebox{0.80}{0.398}&\scalebox{0.80}{0.356}&\scalebox{0.80}{0.385}&\scalebox{0.80}{0.699}&\scalebox{0.80}{0.557}&\scalebox{0.80}{0.395}&\secondres{\scalebox{0.80}{0.408}}&\scalebox{0.80}{0.448}&\scalebox{0.80}{0.452}&\scalebox{0.80}{0.481}&\scalebox{0.80}{0.456}&\scalebox{0.80}{0.588}&\scalebox{0.80}{0.517}&\scalebox{0.80}{0.961}&\scalebox{0.80}{0.733}\\
    \midrule

    \scalebox{1}{ETTm2}  &\boldres{\scalebox{0.80}{0.240}}&\boldres{\scalebox{0.80}{0.319}}&\secondres{\scalebox{0.80}{0.264}}&\secondres{\scalebox{0.80}{0.320}}&\scalebox{0.80}{0.297}&\scalebox{0.80}{0.347}&\scalebox{0.80}{0.255}&\scalebox{0.80}{0.326}&\scalebox{0.80}{0.314}&\scalebox{0.80}{0.365}&\scalebox{0.80}{0.326}&\scalebox{0.80}{0.361}&\scalebox{0.80}{0.283}&\scalebox{0.80}{0.327}&\scalebox{0.80}{0.304}&\scalebox{0.80}{0.349}&\scalebox{0.80}{0.306}&\scalebox{0.80}{0.347}&\scalebox{0.80}{0.324}&\scalebox{0.80}{0.368}&\scalebox{0.80}{1.410}&\scalebox{0.80}{0.823}\\
    \bottomrule
  \end{tabular}
   \end{small}
  \end{threeparttable}
\end{table*}

\subsection{Main Results}
As shown in Table~\ref{tab:long_term_forecasting_results}, our method achieves consistent state-of-the-art performance across all 9 benchmarks, outperforming 10 advanced baselines. Particularly noteworthy is that compared to the second-best method, our approach achieved a 14\% increase in Electricity, 32\% in Solar-Energy, 18\% in Traffic, and 26\% in Workload datasets, highlighting the superiority of our method on complex datasets. Furthermore, we achieved the best performance even on datasets with low forecastability such as ETT and Solar-Energy datasets. We also calculated the standard deviation and conducted statistical significance tests on all datasets in Table~\ref{tab:errorbar}, and achieved the best performance with a confidence level of over 95\% (over 99\% in most cases).

\begin{table}[t]
\vspace{-3mm}
  \caption{Standard deviation and statistical tests for our method and second-best method (TimeMAE) on all benchmarks. We repeat each experiment three times with different random seeds.}\label{tab:errorbar}
  \vspace{-3mm}
  \centering
  \begin{threeparttable}
  \begin{small}
  \renewcommand{\multirowsetup}{\centering}
  \setlength{\tabcolsep}{5pt}
  \begin{tabular}{c|cc|cc|c}
    \toprule
    Model & \multicolumn{2}{c|}{Ours} & \multicolumn{2}{c|}{TimeMAE} & Confidence \\
    Dataset & MSE & MAE & MSE & MAE& interval\\
    \midrule
    Weather & 0.010 & 0.009	& 0.012 & 0.011 & \textbf{99\%}\\
    Solar-Energy & 0.031 & 0.022 &	0.020 & 0.018 &\textbf{99\%}\\
    Electricity & 0.017 & 0.006	& 0.012 & 0.015 & \textbf{95\%}\\
    Traffic & 0.015 & 0.013 & 	0.008 & 0.002 & \textbf{99\%}\\
    workload & 0.021 & 0.023 & 	0.038 & 0.032 & \textbf{99\%}\\
    ETTh1 & 0.022 & 0.025 &	0.031 & 0.051 & \textbf{95\%}\\
    ETTh2 & 0.018 & 0.017 & 0.015 &  0.023 & \textbf{99\%}\\
    ETTm1 & 0.033 & 0.016 &0.002 & 0.029  &\textbf{99\%}\\
    ETTm2 & 0.021  & 0.043	&0.032 & 0.022 & \textbf{95\%}\\
    \bottomrule
  \end{tabular}
    \end{small}
  \end{threeparttable}
\end{table}

\subsection{Model Analysis}

\paragraph{\textbf{Ablations.}}
To confirm the effectiveness of each component in our approach, we conducted detailed ablations on every possible design within the Multiscale TS representation, FusionAttention, and Conditional NF modules. The following findings can be observed from Table~\ref{tab:abla} above:

\begin{table}[htbp]
  \caption{Ablations on each component in predict 96-720 setting of ETTm1. 
  }\label{tab:abla}
  \vspace{-3mm}
  \vskip 0.05in
  \centering
  \begin{threeparttable}
  \begin{small}
  \renewcommand{\multirowsetup}{\centering}
  \setlength{\tabcolsep}{5pt}
  \begin{tabular}{c|cccc}
    \toprule
   Predict Length & \multirow{2}{*}{92} & \multirow{2}{*}{192} & \multirow{2}{*}{336}	& \multirow{2}{*}{720}\\
    Components\\
    \midrule
    Ours &\textbf{0.297} &\textbf{0.340} & \textbf{0.343} & \textbf{0.348}\\
    W/O Multiscale Repr & 0.298 & 0.350 & 0.367 & 0.430\\
    W/O FusionAttention& 0.301 & 0.360 & 0.368 & 0.432\\
    W/O Conditional NF & 0.296 & 0.350 & 0.355 & 0.415\\
    \bottomrule
  \end{tabular}
    \end{small}
  \end{threeparttable}
  \vspace{-4mm}
\end{table}

\begin{itemize}
 \item The exclusion of the Multiscale TS representation module during the ablation study resulted in a notable decline in performance for longer prediction horizons, particularly for prediction horizons of 336 and 720. This finding shows the effectiveness of our proposed method for representing long-term historical information in time series, precisely the essential multi-periodic and long-term dependent characteristics inherent in long-term histories that are crucial for accurate long-term forecasting.

\item The exclusion of the FusionAttention module led to a reduction in performance for all prediction horizons, underscoring the critical role of properly fusing long-term representations and near-term observations. The findings indicate that accurate long-term forecasting cannot be achieved without an appropriate fusion mechanism, even with long-term historical representations.

\item The exclusion of the Conditional NF module decreased the performance, emphasizing the crucial challenge posed by the widespread non-Gaussia/non-linear characteristics present in time series forecasting. This effect was particularly evident in the ablation study, where the performance degradation became more pronounced with an increase in prediction length. These findings highlight the significance of characterizing non-Gaussian/non-linear to achieve accurate long-term forecasting.
\end{itemize}

\begin{figure}[t]
  \centering
    \subfigure{
        \begin{minipage}[t]{0.5\linewidth}
\centering
\includegraphics[width=\linewidth,height=0.55\linewidth]{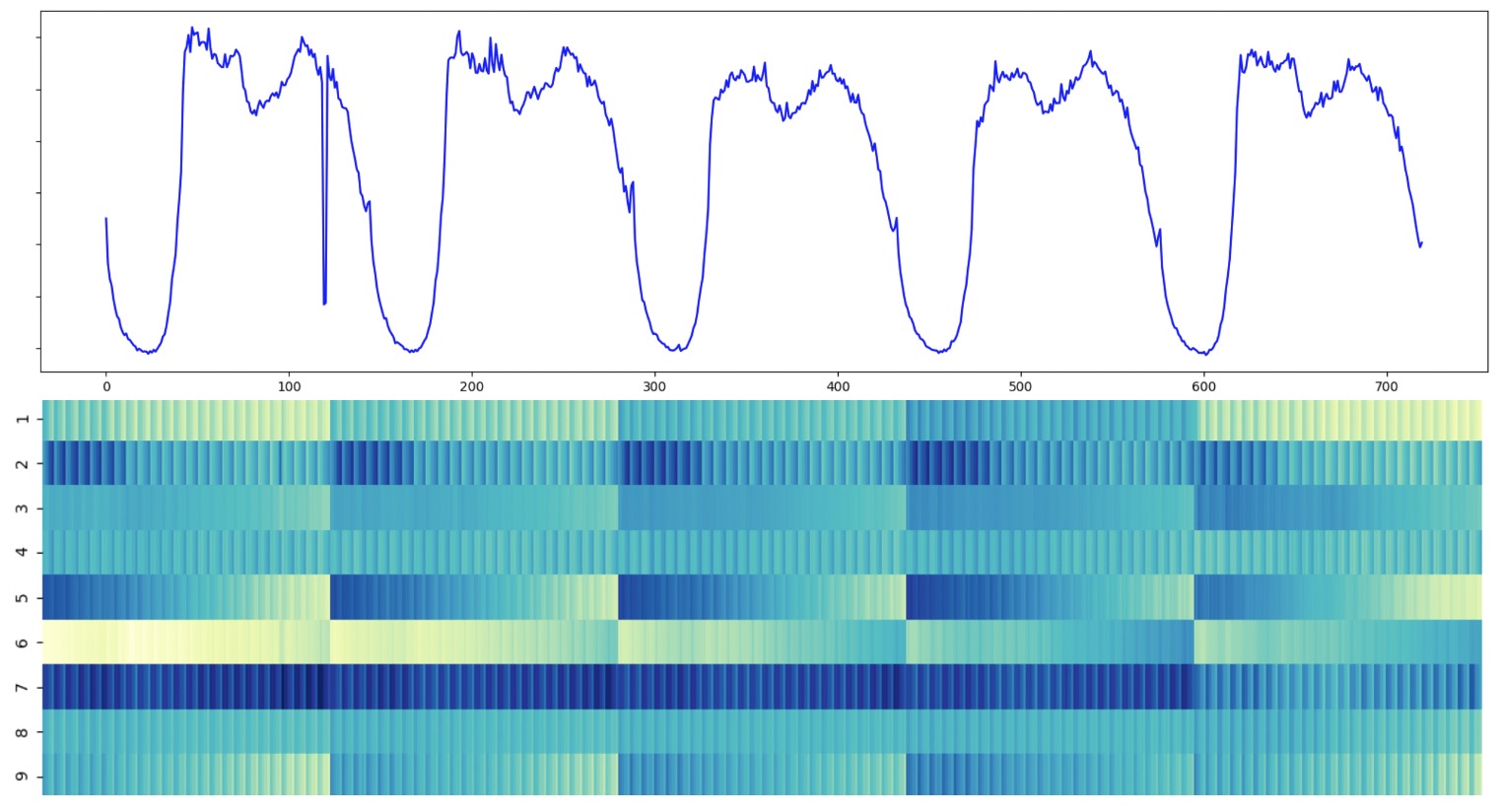}
\end{minipage}%
}%
\subfigure{
\begin{minipage}[t]{0.5\linewidth}
\centering
\includegraphics[width=\linewidth,height=0.55\linewidth]{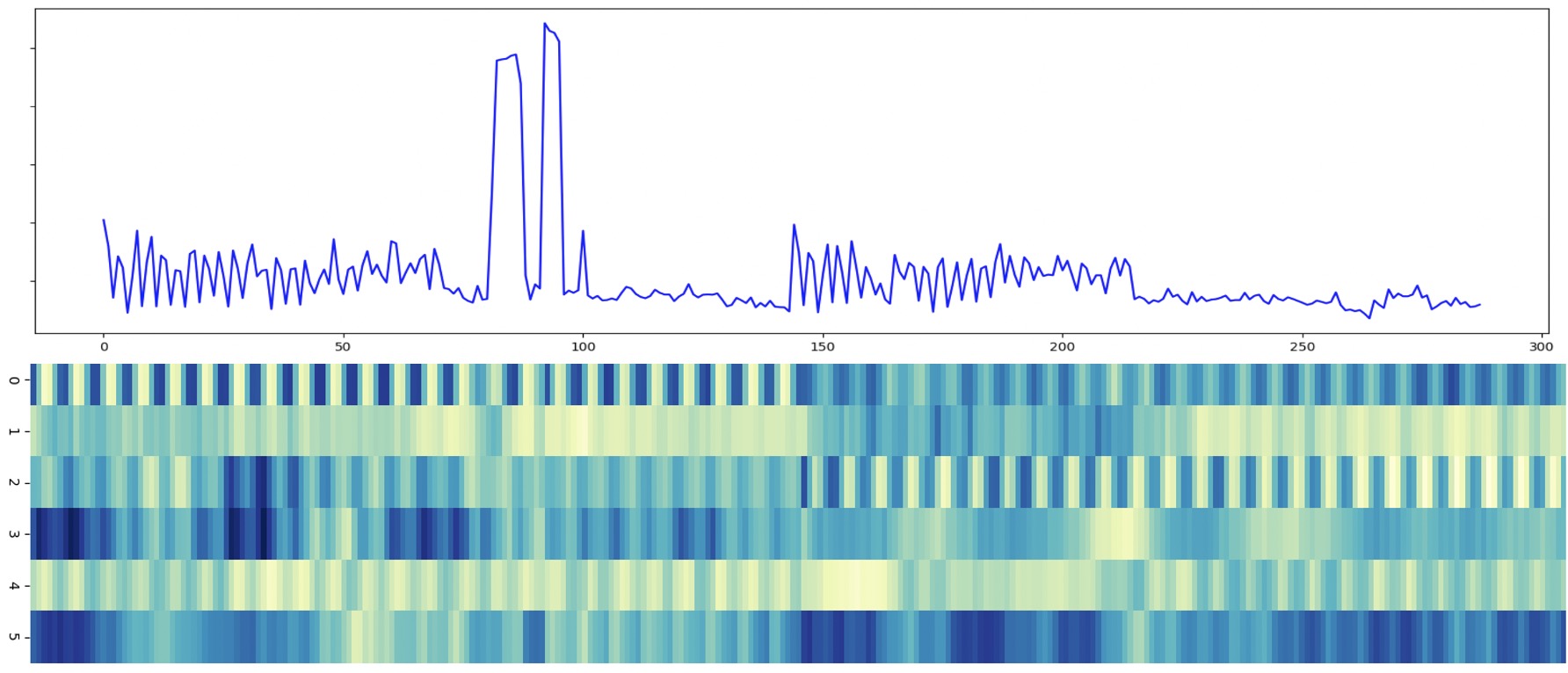}
\end{minipage}%
}%
\vspace{-2mm}
\subfigure{
\begin{minipage}[t]{0.5\linewidth}
\centering
\includegraphics[width=\linewidth,height=0.55\linewidth]{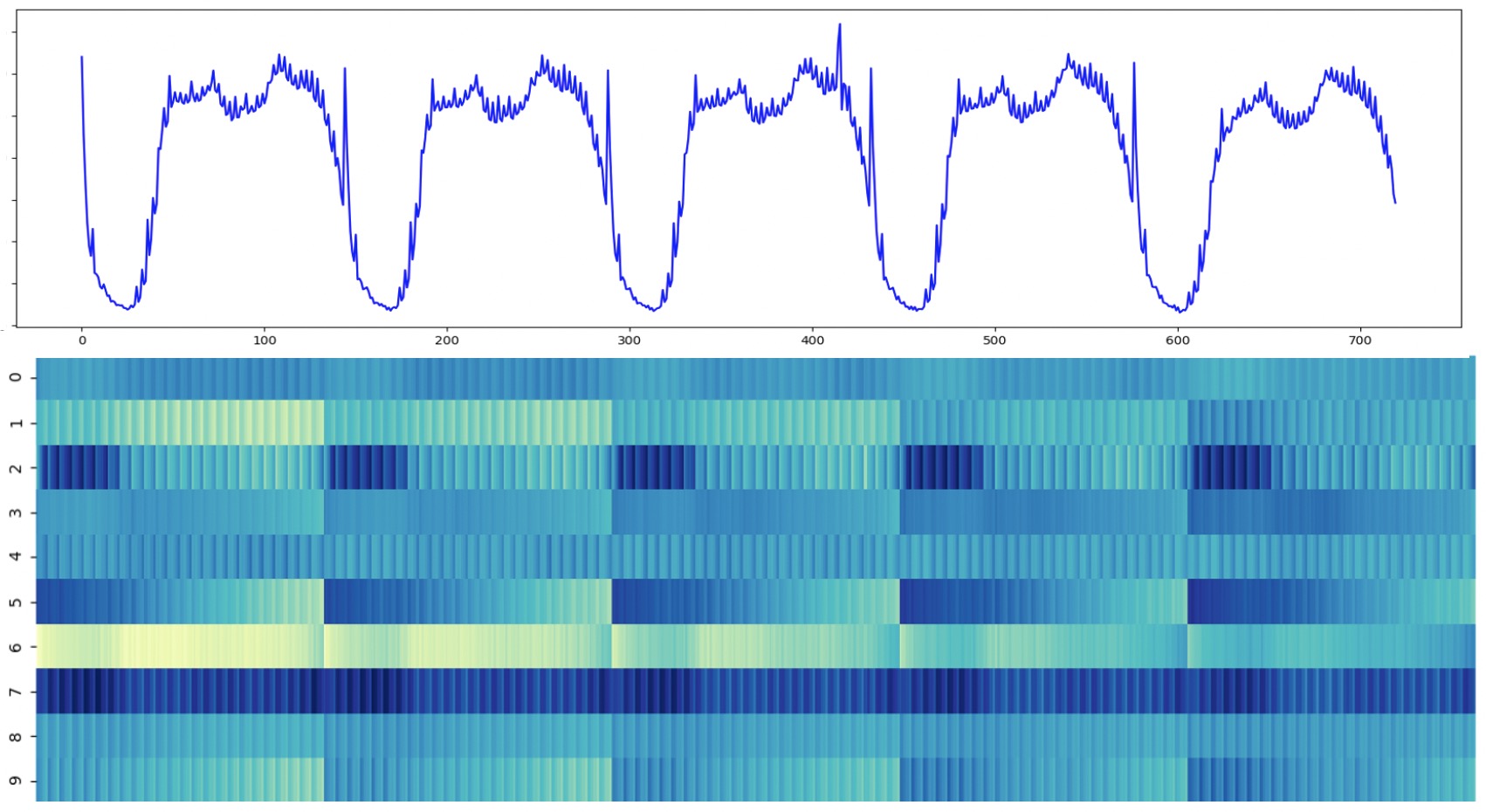}
\end{minipage}%
}%
\subfigure{
\begin{minipage}[t]{0.5\linewidth}
\centering
\includegraphics[width=\linewidth,height=0.55\linewidth]{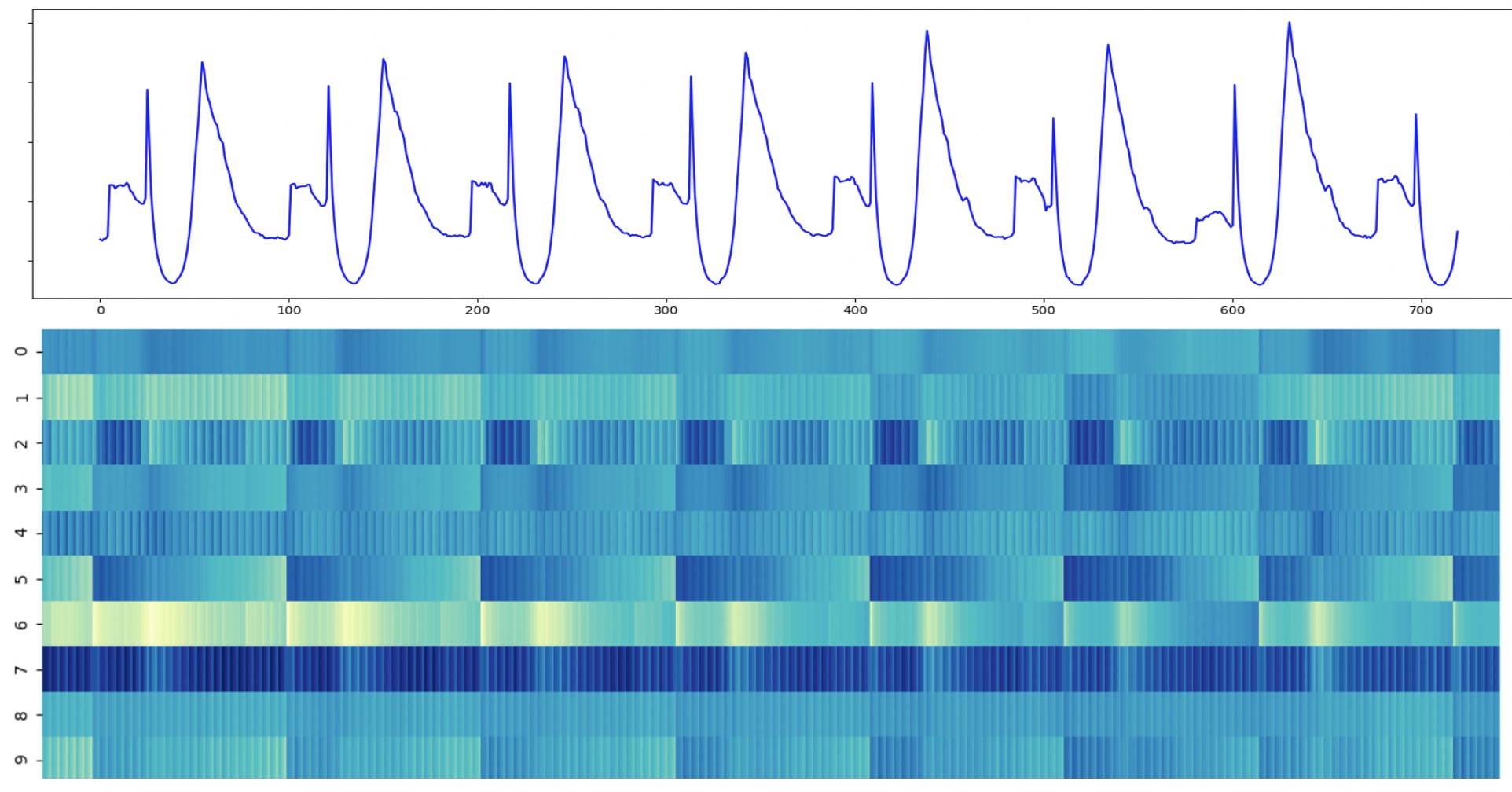}
\end{minipage}%
}
\vspace{-6mm}
  \caption{Visualization of TS representation of different windows in Workload TS.}
  \vspace{-4mm}
  \label{fig:repr}
\end{figure}

\begin{figure}[h!]
  \centering
  \includegraphics[width=\linewidth]{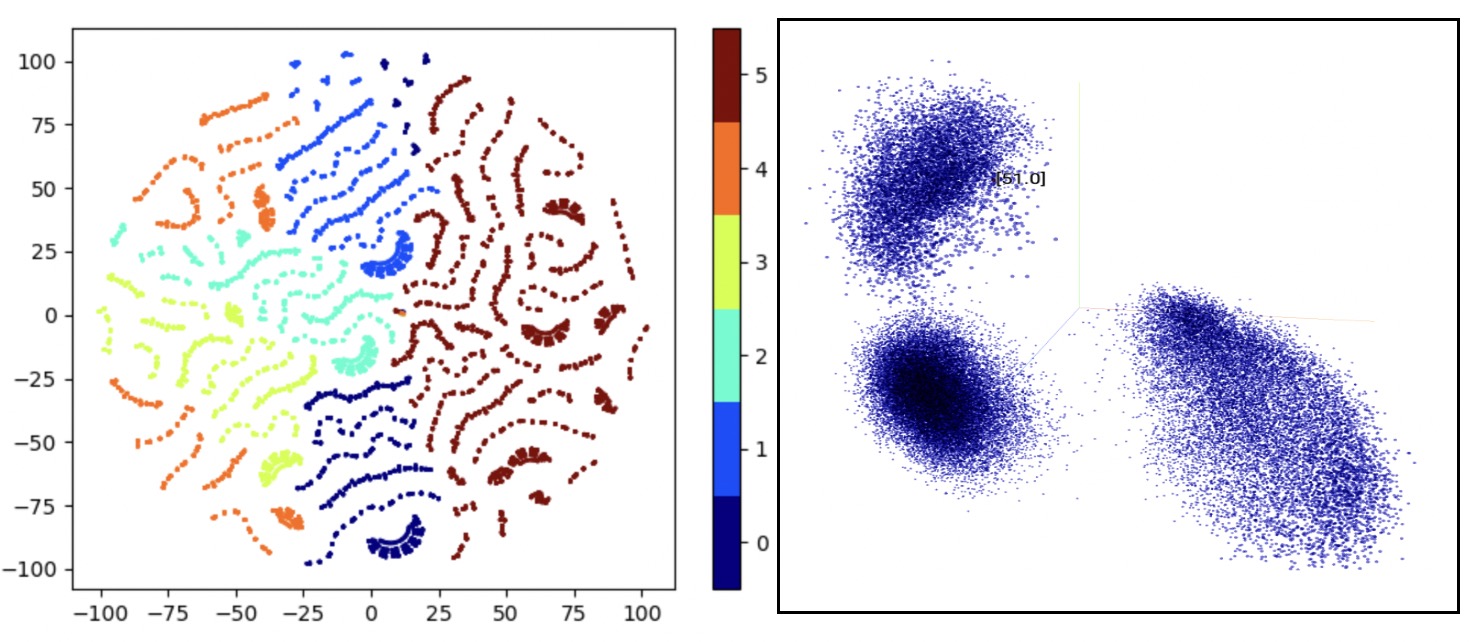}
  \vspace{-6mm}
  \caption{Visualization the clusters of Workload TS representation.}
  \label{fig:cluster}
  \vspace{-3mm}
\end{figure}

\paragraph{\textbf{Representation analysis.}}
In order to facilitate a better analysis of the Multiscale TS representation, we employ visualizations to provide an intuitive understanding of the TS representation. As seen in Figure~\ref{fig:repr}, our TS representation successfully characterizes long-term temporal dependencies across different periods.  We can intuitively see the dependencies between the weekly periods from the representation. Moreover, the TS representation also mines complex nested periods, e.g., the weekly period contains the daily period as shown in the figure. In Figure~\ref{fig:repr}, we plot the visualization of the workload TS representations of different windows, and we can intuitively observe the changes in periods and trends from these representations. Even for TS which does not possess any observable periodic characteristic, our representation can still reveal its corresponding pattern of variations. This attests to the efficacy of our TS representation in capturing long-term temporal dependencies across time periods. Furthermore, we conducted a visual analysis of the clustering of representations, as shown in Figure~\ref{fig:cluster}. 
We can observe that our representations are capable of distinguishing different types of time series in detail. This is evident in both the two-dimensional and three-dimensional clustering visualizations, which demonstrate the effectiveness of our multiscale TS representation.

\begin{figure}[t]
  \centering
  \includegraphics[width=0.9\linewidth]{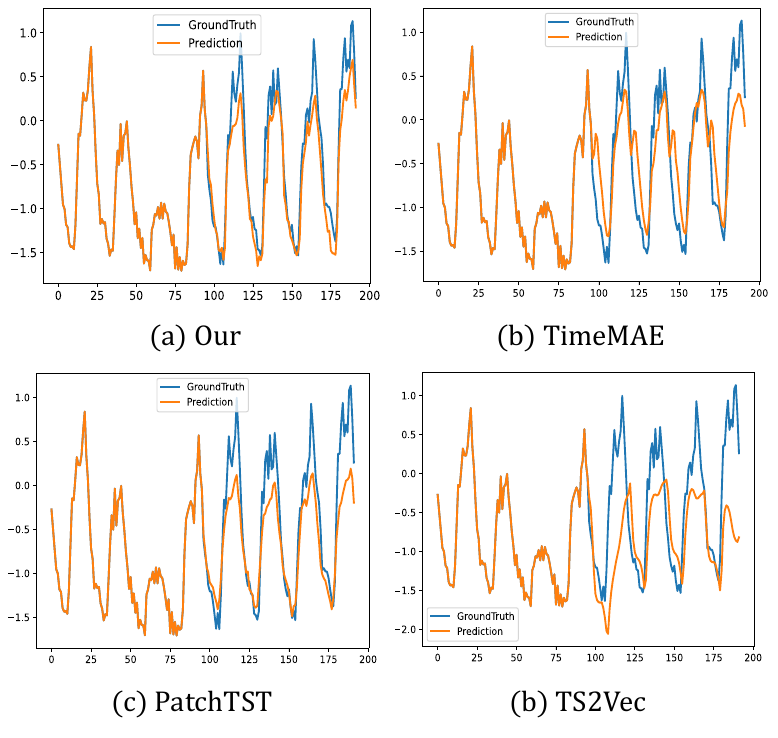}
  \vspace{-3mm}
  \caption{Prediction cases from workload dataset by different models under the input-96-predict-96 settings.}
  \vspace{-3mm}
  \label{fig:showcases}
\end{figure}

\paragraph{\textbf{Forecast results showcases.}}
To evaluate the prediction of different models, we plot the last dimension of forecasting results that are from the test set of the workload TS dataset for qualitative comparison in Figure~\ref{fig:showcases}. Among the various models, our method exhibits superior performance.


\vspace{-3mm}

\section{Application}
Alipay, a leading global mobile payment company, has established its presence in various domains including digital life and digital finance \cite{chu2022hierarchical}. With an expansive cloud computing infrastructure supporting countless microservices, efficient allocation of cloud resources is critical for Alipay's cluster management. Alipay relies on its proprietary predictive autoscaling technology as the fundamental component of its cloud resource management system. The accurate prediction of future workloads for each microservice, specifically in terms of requests per second (RPS), is crucial for the effective scaling of server resources. Presently, our proposed approach has been extensively deployed in Alipay's production cloud environment, achieving remarkable results. The outcomes, as demonstrated in Table~\ref{tab:long_term_forecasting_results}, indicate a significant improvement of 26\% in comparison to the state-of-the-art (SOTA) methods in the real-world environment. It demonstrates the utilization of our autoscaling method based on workload forecasting, which enables the Alipay cloud resource management system to anticipate changes in future workloads through long-term forecasting when traffic fluctuates. This facilitates the scaling of computing resources to allocate those appropriate for the current workload. Upon activating our prediction technique, the number of pods (containers that host microservices) utilized by the microservices has significantly decreased from 1500 to less than 500 (Reduced resource consumption by 67\%), leading to a substantial improvement in resource utilization efficiency.

\section{Conclusion}

This paper presents a novel framework for accurate long-term workload forecasting by incorporating a multiscale time series representation method and a temporal flow fusion model. The framework uniquely captures both long-term historical patterns and near-term observations in the workload time series for superior predictions. To our knowledge, this is the first deep integration of multiscale representation learning and deep forecast modeling for time series. Comprehensive evaluations on 9 benchmarks demonstrate consistent state-of-the-art performance over 10 advanced baselines. Furthermore, when deployed on a real-world cloud resource management system with over a thousand sets of microservices, significant improvements in scheduling and resource management are achieved. This end-to-end framework successfully leverages multiscale representations and temporal fusion to advance the capability of AI systems for long-term time series forecasting.

\bibliographystyle{ACM-Reference-Format}
\bibliography{sample-base}

\end{document}